%% file: main.tex
\documentclass[10pt, a4paper]{article}

\usepackage{natbib}
\usepackage[final]{lrec2026} %

\renewcommand{\abstractname}{Abstract}

\usepackage{tikz}
\usetikzlibrary{positioning}
\usepackage{graphicx}
\usepackage{booktabs}
\usepackage{array}
\usepackage{tabularx}   %
\usepackage{caption}
\usepackage[table]{xcolor}
\usepackage{subcaption}

\usepackage{siunitx}
\usepackage{ragged2e}
\newcolumntype{Y}{>{\RaggedRight\arraybackslash}X}

\usepackage{tikz}  
\usetikzlibrary{arrows.meta,shapes.geometric,positioning}

\usepackage{amsmath,amssymb}

\title{Developing a Guideline for the Labovian-Structural Analysis of Oral Narratives in Japanese}

\name{Amane Watahiki${}^{1,2}$, Tomoki Doi${}^{1,2}$, Akari Kikuchi${}^{3}$, Hiroshi Ohata${}^{3}$\\ {\bf \large Yuki I. Nakata${}^{4}$, Takuya Niikawa${}^{3}$, Taiga Shinozaki${}^{5}$, Hitomi Yanaka${}^{1,2,3}$}} 

\address{
${}^{1}$The University of Tokyo, Tokyo, Japan\\
${}^{2}$Riken, Tokyo, Japan\\
${}^{3}$Kobe University, Hyogo, Japan\\
${}^{4}$Ritsumeikan University, Kyoto, Japan\\
${}^{5}$Keio University, Tokyo, Japan\\
${}^{6}$Tohoku University, Miyagi, Japan \\
amanew@g.ecc.u-tokyo.ac.jp, \{doi-tomoki701, hyanaka\}@is.s.u-tokyo.ac.jp, \\kiku.omu0426@gmail.com, hohata0215@gmail.com, \\dj.y.nakata@gmail.com, niitaku11@gmail.com \\}

\abstract{
Narrative analysis is a cornerstone of qualitative research. 
One leading approach is the Labovian model, but its application is labor-intensive, requiring a holistic, recursive interpretive process that moves back and forth between individual parts of the transcript and the transcript as a whole. 
Existing Labovian datasets are available only in English, which differs markedly from Japanese in terms of grammar and discourse conventions.
To address this gap, we introduce the first systematic guidelines for Labovian narrative analysis of Japanese narrative data.
Our guidelines retain all six Labovian categories and extend the framework by providing explicit rules for clause segmentation tailored to Japanese constructions.
In addition, our guidelines cover a broader range of clause types and narrative types.
Using these guidelines, annotators achieved high agreement in clause segmentation (Fleiss’ $\kappa = 0.80$) and moderate agreement in two structural classification tasks (Krippendorff’s $\alpha = 0.41$ and $0.45$, respectively), one of which is slightly higher than that found in prior work despite the use of finer-grained distinctions.
This paper describes the Labovian model, the proposed guidelines, the annotation process, and their utility.
It concludes by discussing the challenges encountered during the annotation process and the prospects for developing a larger dataset for structural narrative analysis in Japanese qualitative research.
 \\ \newline  }

\begin{document}
\maketitleabstract
\section{Introduction}

\input{tables/dataset_example}

Narrative analysis is central to qualitative research, where thematic and structural approaches are commonly distinguished: the former focuses on what is told, and the latter on how it is told.
Catherine K. Riessman, a leading figure in narrative studies, ``see[s] thematic and structural approaches as the basic building blocks'' of narrative analysis \citep{Riessman:2008}.

Among structural approaches, the \textbf{Labovian model} \citep{LabovWaletzky:1967} offers a simple yet expressive framework that segments narratives into clauses and assigns each clause to one of six discourse functions. \citet{Riessman:2008} characterizes this model as the ``touchstone'' of narrative research because it illuminates not only what happened but also what the event \emph{meant} to the teller. For example, using the Labovian model, she showed that three participants who cited their spouses' affairs as the reason for divorce in a thematic analysis actually conveyed very different meanings in a structural analysis. She concluded that [i]nfidelity was not an objective event, but a phenomenologically different experience'' \citep{Riessman:1989,Riessman:2008}.

Despite their significance, structural analysis and the Labovian approach lack adequate annotation tools and support for (semi-)automation. This contrasts with thematic analysis, which benefits from qualitative data analysis (QDA) software such as NVivo\footnote{\url{https://lumivero.com/products/nvivo/}}, MAXQDA\footnote{\url{https://www.lightstone.co.jp/maxqda/}}, ATLAS.ti\footnote{\url{https://atlasti.com/}}, and Prodigy\footnote{\url{https://prodi.gy/}}, all of which offer auto-coding and visualization features. Although it may, in principle, be possible to apply the Labovian model using these tools, clause segmentation, clause-by-clause tagging, and coding require a holistic interpretation of the data, making it difficult for existing tools to alleviate the labor-intensive procedures inherent in Labovian analysis.

Several corpora in computational linguistics have applied Labov’s framework to written and spoken English narratives~\citep{swanson-etal-2014-identifying, Rahimtoroghi:2014,ouyang-mckeown-2014-towards,SaldiasRoy:2020, Wasserscheidt2021-rb, levi-etal-2022-detecting}. However, prior work is limited in three respects: it often collapses Labov’s six categories into three, largely ignores micro-structure, and rarely makes annotation guidelines public.

In contrast to English and other European languages, there are no reproducible guidelines for Japanese narrative data. \citet{Kodama:2000} made several suggestions for applying this model to Japanese, but since then, no subsequent study has produced a systematically annotated dataset or developed reproducible guidelines for applying the Labovian categories to Japanese interview data.

To fill this gap, our study presents the first reproducible and systematically applicable \textbf{guidelines} for annotating Japanese narratives based on the Labovian model. Our framework introduces a \textbf{classification schema faithful to Labov’s original six categories} while extending the model to incorporate micro-level structure and a broader range of narrative types. Using these guidelines, we constructed a pilot dataset of dementia carer interviews annotated for clause segmentation, narrative span identification, narrative type detection, and micro- and macro-level structural-function classification. Agreement among the three annotators on clause segmentation was high (Fleiss’s $\kappa = 0.80$). At the same time, agreement on micro- and macro-level classification was moderate (Fleiss’s $\kappa = 0.40$ and $0.45$, respectively), with the latter still slightly higher than that reported in previous studies despite the use of finer-grained distinctions. An example of the annotated data, modified to protect participant privacy, is shown in Table~\ref{tab:dataset-example}. These results demonstrate that our guidelines\footnote{\url{https://github.com/ynklab/Labov-guideline.git}} provide a foundation for creating larger, publicly shareable datasets for structural narrative analysis in Japanese. While we focus on constructing a dataset based on the Labovian framework, this framework is also expected to contribute to research on the well-being of family carers of people with dementia.

\section{The Labovian Model of Narrative}

\subsection{Definition of Narrative}

William Labov is an American linguist widely regarded as the founder of variationist sociolinguistics. In their influential article ``Narrative Analysis: Oral Versions of Personal Experience'' \citep{LabovWaletzky:1967}, Labov and Joshua Waletzky proposed a theoretical framework for identifying the structure of oral narratives of personal experience. \citet{Riessman:2008} notes that Labov's model represents one of the two major types of ``structural analysis'' in narrative research.

According to Labov, a ``narrative'' (which corresponds to what we refer to as a ``story'' in this paper) is a particular way of retelling past events. A narrative matches the order of the independent clauses with the original events referred to'' \citep{Labov:2013}. In this sense, a ``narrative'' recounts a sequence of specific past events and therefore necessarily includes at least two clauses, each representing a distinct event. In Labov’s framework, clauses representing distinct events are referred to as ``narrative clauses'' (see Table~\ref{tab:dataset-example} for an example).

\subsection{Micro- and Macro-Structural Functions}
\label{subsec:labov-model}
However, a narrative composed solely of narrative clauses constitutes merely a report of a sequence of events that is monotonous and lacks background or interpretive information. When narrating a story in conversation, the speaker must claim an extended turn and convince the listener that the account is worth hearing. Unless the event itself is extraordinarily unusual and occurs in an exceptional situation in which the narrator can reasonably expect it to be accepted by the listener, simple reporting of events is rarely sufficient to accomplish these interactional goals. Thus, in addition to reporting that certain events occurred through narrative clauses, the narrator performs other kinds of discourse work within the story. Labov and Waletzky identified such clauses as \textit{free} and \textit{restricted} clauses (see Section~\ref{subsec:micro}), which convey information with temporal characteristics different from those of narrative clauses (see the examples in Table~\ref{tab:dataset-example}). The information provided by free or restricted clauses is not tied to specific events. Instead, it describes states or conditions that hold either throughout the narrative or within limited portions of the narrative time frame.

Narrative, free, and restricted clauses thus constitute a classification based on the temporal character of the information they convey, forming what may be called the ``micro-structure'' of a narrative. However, a further question arises: for what purpose does the narrator use these free and restricted clauses? Labov addressed this question by examining how each clause serves different functions within the narrative as a whole. At this ``macro-structural'' level, he identified six primary functions—\textit{Abstract}, \textit{Orientation}, \textit{Complication}, \textit{Evaluation}, \textit{Resolution}, and \textit{Coda}—which describe how narrators frame, organize, interpret, and close their stories (see Section~\ref{subsec:macro}).

For analysts, identifying micro-level clause types makes it easier to determine which parts of a narrative describe the sequence of events and which reflect the speaker’s other discourse activities, which in turn helps them identify the macro-level function of each clause. In doing so, analysts can visualize what narrators are doing in each clause and how they construct the overall structure of their stories, thereby enabling researchers to access the meanings that narrated events hold for the narrator.

\subsection{Extensions by Riessman: Habitual and Hypothetical Narratives}
\label{subsec:extension-by-riessman}
There are other ``genres'' of narratives than those which Labov deals with. \citet{Riessman:1990}, for example, included \textit{habitual narratives} and \textit{hypothetical narratives} for her analysis. Since the interview data from dementia-carers include these kinds of narratives, especially habitual ones, we decided to develop a guideline to apply to these other types of narratives. Accordingly, ``narrative'' in Labov's definition is called ``story'' in our paper, following  \citet{Riessman:1990}.

\citet{Riessman:1990} explains these narrative types as follows. A \textit{habitual narrative} tells of the general course of events over time, rather than what happened at a specific point in the past'' \citep{Riessman:1990}. For example, the following constitutes a minimal habitual narrative \citep{Labov:2013}:
\begin{enumerate}
    \item He would hit me.
    \item I'd hit him back. 
\end{enumerate} 
In contrast, a \textit{hypothetical narrative} is ``a narrative about events that did not happen, capped off by a story (about events that did happen)'' \citep{Riessman:1990}. Such narratives describe imagined or counterfactual sequences, often employing modal or conditional constructions to explore alternative possibilities. Labeling clauses alone does not automatically yield interpretive insights into the data. However, as \citet{Riessman:2008} emphasizes, adopting Labov’s framework offers a substantial advantage at the initial stage of analysis by providing a principled foundation for exploring how narrators structure experience, assign meaning, and position themselves through storytelling.

\section{Related Work}

\subsection{Narrative Corpora and Annotation Schemes}
In computational linguistics, several corpora have been developed that apply Labov's categories to written and spoken data ~\citep{swanson-etal-2014-identifying, ouyang-mckeown-2014-towards,SaldiasRoy:2020,Wasserscheidt2021-rb, levi-etal-2022-detecting}. However, (i) they tend to collapse Labov's six-way scheme into three, (ii) ignore micro-level classification, or (iii) do not publicize the annotation guideline. 

Exceptions to the first limitation include \citet{ouyang-mckeown-2014-towards} and \citet{Wasserscheidt2021-rb}, which preserve all six Labovian categories. However, neither study provides a fixed, reproducible annotation guideline. 
\citet{SaldiasRoy:2020} released annotation guidelines based on 594 spoken narratives and over 10,000 annotated clauses, but simplified the framework to three categories and omitted micro-level labels and narrative-type distinctions. By contrast, our scheme retains all six macro-level categories, adds micro-level classification, and distinguishes among narrative types.

\subsection{Discourse Frameworks and Labovian Annotation Schema}

A wide range of discourse frameworks—such as RST \citep{MannThompson:1988}, SDRT \citep{LascaridesAsher:2007}, and PDTB \citep{PennDiscourse:2008}—have been proposed for discourse analysis. However, these annotation schemes tend to be overly detailed, which can hinder inter-annotator agreement and make them less practical for large-scale narrative annotation. 
Consistent with this view, \citet{Rahimtoroghi:2014} compared prominent discourse annotation schemes such as RST and PDTB with simpler annotation schemes, including the Labovian framework, and concluded that simpler schemes ``enable the quick labeling of large amounts of narrative text with reduced annotation labor costs.'' Therefore, rather than adopting detailed and complex discourse frameworks such as RST, we employed a simpler yet sociolinguistically grounded and expressive annotation schema based on the Labovian model.

\subsection{Japanese Narrative Studies}
The Labovian model exhibits certain affinities with the discourse analysis frameworks discussed above. One of the most salient parallels lies in the correspondence between the Labovian narrative clause and the \textit{Narration} discourse relation in SDRT. Several Japanese discourse relation corpora have also been developed~\citep{kishimoto-etal-2018-improving, kubota-etal-2024-annotation-japanese}. However, the relationship between these two models has not yet been systematically investigated. To facilitate such an investigation, it would be beneficial to develop reliable annotation guidelines for the Labovian model.

Applications of the Labovian model to Japanese remain limited. While most studies in Japan drawing on the Labovian model have analyzed the role of specific grammatical expressions (e.g., negation and conjunctions) in narratives, \citet{Kodama:2000} examined the feasibility of applying Labov’s framework to Japanese oral narratives, identifying key challenges such as the embedding of subordinate and quoted clauses, the complexity of verb forms and clause-linking particles, and the indeterminacy of sentence boundaries in spoken discourse. Despite these contributions, existing studies have not yet constructed clause-level annotated corpora or developed systematic annotation schemes based on Labov’s model.

To advance this line of inquiry, our study develops annotation guidelines for structural narrative analysis in Japanese that are faithful to, and extend, Labov’s original model. Rather than presenting the dataset itself as the primary contribution, we emphasize the methodological framework that these guidelines provide for the future development of larger and more consistent datasets for Japanese narratives.

\section{Dataset and guideline}
In this section, we describe the dataset characteristics (\ref{subsec:dataset-characterizaion}), data collection (\ref{subsec:data-collection}), the annotation pipeline (\ref{subsec:dataset-annotation}), how each annotation task was carried out, including the challenges we encountered and how we addressed them (\ref{subsec:clause-segmentation}–\ref{subsec:macro}), and the agreement metrics (\ref{subsec:agreement-and-metrics}).

\subsection{Dataset Characterization}
\label{subsec:dataset-characterizaion}
Our corpus contains 965 clauses derived from sixteen interviews involving two speakers: the interviewer and the interviewee. 
Speaker contributions are highly skewed toward interviewees (847 clauses) relative to interviewers (118 clauses), with interviewer speech consisting mainly of questions, short prompts, and backchannels.

In NLP research, ``narrative'' is commonly divided into two categories: written narrative (e.g., novels and movie scripts) and oral narrative (e.g., interviews and social media posts)~\citep{doi_yanaka_2024_narrative_nlp}. As noted above, the narratives in our dataset belong to the latter category. In contrast to news articles and general social media posts, which are often written in the third person or focus on current events, caregivers’ narratives are distinct in that they are typically told in the first person and tend to focus on the speaker’s past experiences. However, our annotation framework does not depend on these distinctive features of caregivers’ narratives.

From each interview, we extracted the interviewees’ responses to questions about (i) when they experience happiness or hardship in their current situation and (ii) the challenges they face. We selected these passages because they tend to contain more narrative content than other parts of the transcripts. Accordingly, the unit of annotation is an interview segment in which the interviewee discusses either of these topics.

Three postdoctoral researchers participated in the annotation process, and gold labels were assigned by majority vote. When no two annotators agreed, the final label was determined through discussion. For the micro-level labels, Narrative clauses accounted for 48\% of all annotated clauses (193 instances), Free clauses for 34\% (139 instances), and Restricted clauses for 17\% (68 instances). The distribution of macro-level labels is summarized in Table~\ref{tab:macro-labels-total}.
\input{tables/macro_labels_total}

\subsection{Data Collection}
\label{subsec:data-collection}
Our data were drawn from semi-structured interviews with family caregivers of people with dementia in Japan.

\begin{itemize}
\item \textbf{Format:} The interviews were conducted from October 2024 to March 2025 in Hyogo and Osaka Prefectures, Japan. Each interview lasted approximately 60–90 minutes. The selected interviews were conducted primarily one-on-one, either online or in person, depending on the interviewee’s convenience.
\item \textbf{Participants:} Caregivers were mainly in their fifties and sixties, although younger participants in their thirties and forties, as well as older participants in their eighties, were also represented. They lived in Hyogo and Osaka Prefectures, Japan. Their occupations included homemakers, part-time workers, and care professionals. Participants reported an average of seven years of caregiving experience, with the longest duration extending to two decades.
\item \textbf{Interview questions:} Interviewers were instructed to ask a specific set of questions during each interview, including (i) ``In your current life, when do you experience happiness or hardship?'' and questions about the difficulties and challenges encountered in everyday care.
\item \textbf{Care recipients:} Care recipients represented the full range of care-need levels (1–5) and were mostly co-residing, although some were living in care facilities.
\end{itemize}

All interviews were transcribed and de-identified to remove personal names and locations, and informed consent was obtained from all participants.

\subsection{Dataset Annotation}
\label{subsec:dataset-annotation}
We designed a multi-level annotation pipeline, which was iteratively refined through version~4 of the guidelines. The process involved three postdoctoral annotators with backgrounds in philosophy and psychology, all of whom had experience in qualitative research and worked collaboratively throughout the annotation process.

The key stages of the pipeline were as follows:
\begin{enumerate}
\item Clause segmentation.
\item Narrative type and span detection.
\item Clause-level annotation at two levels: \emph{micro} and \emph{macro}.
\end{enumerate}

In this corpus, interviewer and interviewee utterances are clearly distinguished. The interviewer’s utterances are included as contextual information but are neither segmented nor annotated for narrative structure. For the micro- and macro-level annotations, the two levels were applied simultaneously because each provides cues for the other, and this interdependence reflects how analysts perform the task in practice.

\subsubsection{Clause segmentation}
\label{subsec:clause-segmentation}
\input{tables/narrative_span_length}

Clause segmentation is the foundation of structural annotation. Each clause is defined as representing a distinct state of affairs. Paraphrases (My mo- father is...'') and insertions (My mother — she was a high school teacher — would walk every Sunday…'') are not segmented.

Many parts of our guidelines rely on \citet{Kodama:2000}’s suggestions for handling challenges in Japanese clause segmentation: embedded clauses are not counted as independent clause structures. Quoted speech is not recognized as a separate clause when no quotative marker is present. Modifying clauses are not recognized as independent clause structures.

Despite adopting these principles, our annotation process revealed additional challenges. Although modifying clauses should not be segmented according to \citet{Kodama:2000}‘s proposal, this prescription appears inadequate in some cases where such clauses modify formal nouns such as \textit{toki} (``time/when''), \textit{koro} (``around the time/when''), \textit{baai} (``case/situation''), and \textit{tokoro} (``point/moment/place''). We analyzed our annotation results and found that formal nouns modified by clauses function in three ways: they may modify a main event, function as discourse topics, or serve as arguments. To handle these patterns, we introduced explicit rules in our guidelines for formal nouns, including the following:

\begin{itemize}
\item When such a phrase functions as a discourse topic (e.g., \textit{Watashi ga hajimete Tokyo ni kita toki wa ichiban shiawase datta} ``When I first came to Tokyo, I was the happiest''), it is segmented separately. \item When it serves as a subject or complement (e.g., \textit{Watashi ga hajimete Tokyo ni kita toki ga ichiban shiawase deshita} ``The time when I first came to Tokyo was the happiest moment'') or as a nominal modifier (e.g., \textit{Watashi ga daigaku ni nyuugaku shita toki no omoide wa…} ``The memories of when I entered university…''), it is not segmented.
\end{itemize}

This operationalizes the distinction between \textbf{topic} and \textbf{subject} roles: formal-noun clauses functioning as topics are segmented because they organize discourse structure. In contrast, those functioning as subjects remain unsegmented because they form part of the propositional content.

\subsubsection{Narrative type and span detection}
\label{subsec:narrative-span}

Narrative type and span detection aims to identify contiguous stretches of clauses that together form a coherent narrative unit, as well as the type of narrative represented by each span. Annotators mark the beginning (\texttt{S}) and end (\texttt{E}) of each narrative span, each of which may belong to one of three types: \emph{Story}, \emph{Habitual Narrative}, and \emph{Hypothetical Narrative}. Each span must contain at least two clauses. Table~\ref{tab:narrative-span-length} summarizes these three categories of narrative spans, along with their definitions and distributions in our dataset.

We followed four main criteria for span detection:
\begin{itemize}
\item \textbf{Introductory cues:} To begin telling a narrative, the teller must situate the listener in the specific time and place of the past events they are about to recount. Parts of the transcript where tellers attempt to do this serve as cues for identifying the starting point of a narrative.
\item \textbf{Identity of the event sequence:} A narrative often continues as long as the narrator refers to the same sequence of events. Even if the narrator’s conversational turn is not interrupted, a new narrative begins when the event sequence under discussion changes.
\item \textbf{Role in interaction:} Narratives frequently occur as elaborations in response to the interviewer’s questions. Extended elaborations, rather than brief direct answers, are more likely to be marked as the beginning of a narrative.
\item \textbf{Discourse markers:} Interviewee expressions such as \textit{A, sou da} (Oh, that reminds me'') and \textit{Sono toki wa} (``At that time'') frequently signal narrative onset. In contrast, interviewer responses such as \textit{Sore wa taihen deshita ne} (``That must have been tough'') often indicate the end of a narrative.
\end{itemize}

Note that the interviewer’s utterances are neither segmented nor annotated, although they provide valuable cues for identifying narrative boundaries. When a story ends, the interviewer’s response signals that the narrated episode has been understood by the participants as complete.

The narrative span and type detection task was separated at the final stage of our guideline development process. In the earlier stages, span and type detection were performed simultaneously with micro- and macro-level classification. However, it became apparent that misalignment of span boundaries significantly affected classification results. We therefore decided to separate these two tasks. Furthermore, due to the intrinsic complexity of narrative span detection, inter-annotator agreement remained low: Fleiss’s $\kappa$ for Boundary Similarity was approximately $0.17$ (for this metric, see Section~\ref{subsubsec:clause-segmentation}). Although there is room for improvement through further refinement of the guidelines, the final reference spans in our dataset were determined solely through discussion among the annotators before proceeding to the clause classification stage.

\subsubsection{Clause type classification – Micro level}
\label{subsec:micro}

\begin{figure}[t]
  \centering
  \includegraphics[width=\linewidth]{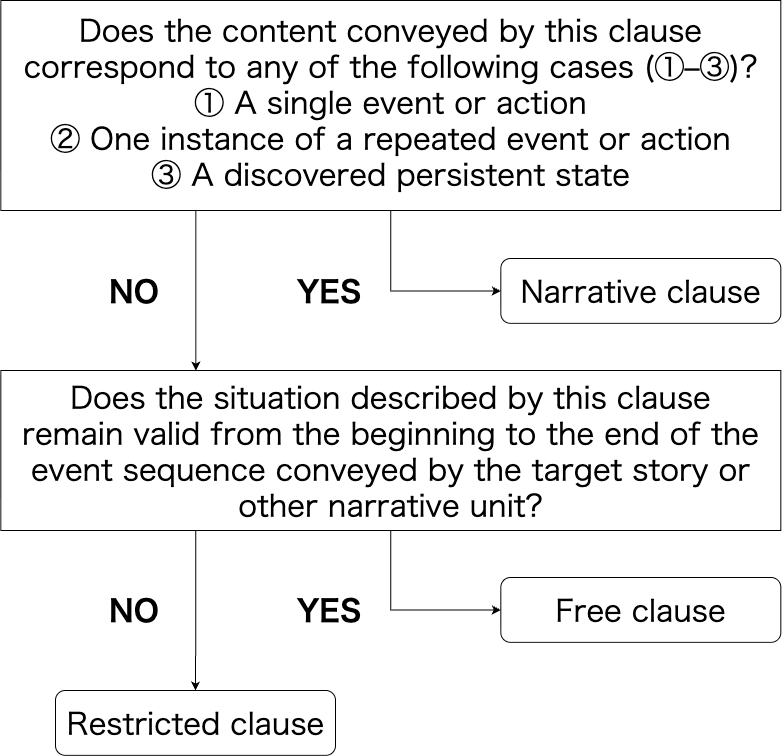}
  \caption{Micro-label determination chart across narrative clause types.}
  \label{fig:micro_label_chart}
\end{figure}

At the micro level, each clause was classified into one of three types—\emph{Narrative}, \emph{Free}, or \emph{Restricted}—all of which presuppose that the described event or state actually occurred. Consequently, we assigned no micro-level labels to clauses in hypothetical narratives.

As discussed in Section~\ref{subsec:labov-model}, a \textit{Narrative clause} describes a specific event or action. In contrast, \textbf{Restricted clauses} denote information that holds only within a limited portion of the narrative, whereas \textbf{Free clauses} provide background information or general statements that remain valid throughout the entire period at issue in the narrative. Adopting this ``temporal criterion,'' we developed a decision chart for micro-level classification (see Figure~\ref{fig:micro_label_chart}).

A key clarification concerns \textbf{Narrative clauses}. 
In addition to discrete events, they can also include the reporting of \emph{newly discovered or encountered enduring states}—that is, states or conditions that existed beforehand but became relevant to the narrator only at a particular point in the narrative. 
For example, consider the sentence: ``When I looked closely at my cat, who has such a cute face, I saw that the pupils were still wide open’’
\footnote{Adapted from an example in \citet{Kodama:2000}.}
Here, the cat's pupils were already dilated, but the narrator discovers and reports this fact only at that moment, making it a \textit{Narrative} clause.
By contrast, the clause ``who has such a cute face'' conveys a general truth not tied to discovery and thus qualifies as a \textit{Free} clause rather than a \textit{Narrative} clause.
This rule operationalizes \citet{Fleischman:1990}'s proposal that a clause should be considered a narrative clause when it is ``essential to plot development.''

\subsubsection{Clause type classification – Macro level}
\label{subsec:macro}

Macro-level labels capture the global discourse functions of clauses within a narrative. Their definitions are provided in Table~\ref{tab:macro-labels-total}; examples are shown in Table~\ref{tab:dataset-example}.

One caveat should be noted regarding our annotation scheme. Labov distinguished between two kinds of \emph{evaluation}: \emph{internal evaluation} and \emph{external evaluation} \citep{LabovWaletzky:1967}. External evaluation occurs when the narrator temporarily suspends the progression of the story to address the listener directly and emphasize the point or significance of the events. For example, an utterance such as It was really terrible'' explicitly signals the narrator's stance and invites the listener's empathy. In contrast, internal evaluation does not interrupt the flow of the narrative; instead, the narrator's attitude is implied in a report of unfolding events through characters' thoughts, reported feelings, or marked grammatical devices such as negation, modality, or hedging. For instance, in I tried not to cry, but I couldn’t help it,’’ the evaluation is embedded in the action rather than stated explicitly. Our scheme focuses only on \emph{external evaluation}, that is, clauses that explicitly pause the narrative to highlight meaning, interpretation, or significance.

Although Labov reserved macro-structural analysis for single-event stories, we extend the framework to \textbf{habitual narratives} consisting of multiple clauses. Our interview data often contained accounts that highlighted a complete cycle of such routines, depicting how the activity unfolded, what motivated it, and what consequences it entailed. Within these cycles, we frequently observed structural patterns analogous to those found in stories. An illustrative example is provided below to show how a habitual narrative can display story-like structure within a single cycle of repetition: After I lost my job, I started exercising obsessively'' (\textit{Orientation}). Every morning, I did radio exercises, ran 10 km, went to the job center, and trained at the gym’’ (\textit{Complication}). Eventually, I damaged my health'' (\textit{Resolution}). Now I hardly exercise at all’’ (\textit{Coda}).

The status of \textbf{hypothetical narratives}, however, remains less clear. Therefore, we did not apply micro- or macro-level labels to hypothetical narratives. Future work will need to examine whether such narratives can assume a structure analogous to that of stories or habitual narratives.

\subsection{Metrics and Agreements}
\label{subsec:agreement-and-metrics}

\subsubsection{Clause segmentation}\label{subsubsec:clause-segmentation}
We evaluated clause segmentation agreement using the edit-based metrics implemented in the \textsf{SegEval} package\footnote{\url{https://segeval.readthedocs.io/en/latest/}}. Following \citet{Fournier:2013}, we report two measures: \textbf{Boundary Edit Distance} (BED) and an inter-coder agreement coefficient, namely Fleiss’ $\kappa$, based on \textbf{Boundary Similarity} (B; \citealp{Fournier:2013}). BED quantifies the minimum number of edit operations—insertions, deletions, and transpositions—required to transform one segmentation into another. B normalizes this cost by the number of potential boundary positions and then weights near-boundary alignments. While BED is an unbounded measure reflecting the number of operations required to align two segmentations, B ranges from $0$ (no agreement) to $1$ (perfect agreement). Higher values of B indicate better agreement, whereas lower values of BED indicate that fewer edits are needed to reconcile two segmentations. \citet{Fournier:2013} proposes a B-based calculation of Fleiss’ $\kappa$, which we adopt as our agreement metric.

Pairwise comparisons among the three annotators were conducted on each interviewee’s response sequence for a single topic (see Section~\ref{subsec:dataset-characterizaion}), yielding a mean Boundary Similarity of $B = 0.80$ and Fleiss’ $\kappa = 0.80$. This value of $\kappa$ is generally considered to indicate substantial agreement~\citep{Pustejovsky2012-xz}. For comparison, \citet{Mildner2024-spontaneous-thought} report a Fleiss’ $\kappa$ of $0.58$, which they characterize as ``moderate,'' for a task that segments transcripts into ``units of thought,'' defined as the minimal units of text that can stand on their own as thoughts. According to the authors, a unit of thought identified by the coders corresponds in many cases to an independent clause and, in some cases, to a dependent clause or partial sentence when it expresses a complete and contentful thought, making this task comparable to our clause segmentation task. The average BED per 100 possible boundaries was $13$.

Following \citet{Fournier2013-uv}, we also report BED and $\kappa$ for two random segmentations. For each fragment corresponding to a single topic within an interview, a random segmentation with the same number of boundaries as the mean human segmentation was generated. The comparison between two random segmentations yielded Fleiss’ $\kappa = 0.30$ and a BED of $85$ per 100 possible boundaries. These results suggest that the guidelines we developed improve inter-annotator agreement, with the remaining differences reflecting only a limited amount of manual correction effort.

\subsubsection{Micro- and  Macro clause type classification}
For narrative type and structural labels, we calculated nominal Krippendorff’s $\alpha$. For micro-level labels (\textit{Narrative}, \textit{Free}, and \textit{Restricted}), Krippendorff’s $\alpha$ was approximately $0.40$, while for macro-level labels (\textit{Abstract}, \textit{Orientation}, etc.), it was approximately $0.45$. These values are comparable to or slightly higher than the value we calculated ($0.31$) from a released English dataset associated with \citet{SaldiasRoy:2020}, in which clause-level annotation was conducted via Amazon Mechanical Turk. Although Saldias and Roy did not report inter-annotator agreement in their paper, we calculated it from their released dataset \citeplanguageresource{SaldiasRoy2020_RTNData}. A direct comparison is not possible because of differences in the experimental settings, and no other benchmark for Labovian annotation is currently available. Nevertheless, our higher agreement suggests that the decision charts and extended guidelines contribute to more consistent labeling, although substantial ambiguity remains. Although this remains a hypothesis, identifying the micro-structure of narratives may lead to higher macro-level classification scores.

We additionally report the exact-match rate, $P(\text{complete agreement} \mid \text{a label was chosen})$, for micro- and macro-level labels (Table~\ref{tab:exact-match}). The results show that \textit{Narrative} clauses at the micro level achieve the highest consistency ($0.52$), while \textit{Complication} clauses at the macro level are identified most reliably ($0.45$).

\textit{Restricted} clauses at the micro level ($0.08$), as well as \textit{Resolution} ($0.10$) and \textit{Coda} ($0.08$) at the macro level, exhibit low exact-match rates, indicating that these categories are more difficult to identify consistently. This may reflect both their relative scarcity in the dataset and the inherent ambiguity involved in distinguishing them from neighboring functions.

\begin{table}[th]
\centering
\small

\begin{minipage}{0.48\linewidth}
\centering
\input{tables/exact-match_micro}
\end{minipage}%
\hfill
\begin{minipage}{0.48\linewidth}
\centering
\input{tables/exact-match_macro}
\end{minipage}
\caption{Exact-match rates for micro and macro labels 
($P(\text{match}\mid\text{chosen})$). Highest values are in bold.}
\label{tab:exact-match}

\end{table}

\section{Discussion}

\subsection{Error Analysis}

\medskip
\noindent\textbf{Segmentation challenges.}~
One challenge specific to spoken interview transcripts involved distinguishing insertions from paraphrases and identifying quoted segments. Annotators often had to rely on contextual cues to infer slashes and quotation marks that were absent from the transcripts, which increased ambiguity in clause segmentation.

\medskip
\noindent\textbf{Restricted clauses.}~
The exact-match results indicate that \emph{Restricted} clauses are the most difficult for annotators to identify consistently (cf.\ Table~\ref{tab:exact-match}). Identifying them requires a precise understanding of the temporal scope of the narrated events and of the relevant period within the overall narrative. The annotation guidelines should be refined to help annotators capture the temporal relationships between events more accurately.

\medskip
\noindent\textbf{Codas, reference time, and narration time.}~
Low inter-annotator agreement was also observed for the macro-level labels \emph{Resolution} and \emph{Coda}. These two labels are closely intertwined, as in many cases it is difficult to determine whether a clause should be classified as \emph{Resolution} or \emph{Coda}. This observation suggests that greater confidence in identifying one of these clause types may, in turn, facilitate the identification of the other.

The classification of \emph{Coda} may be improved by incorporating tense–aspect theory into our guidelines. \citet{Reichenbach:1947} introduces the notion of \emph{reference time} (the time under discussion), distinct from \emph{event time} (the time at which the event takes place) and \emph{speech time} (the time of utterance). With this distinction, \emph{Codas} could be characterized as clauses that shift the reference time forward to the time of utterance. If so, developing a computational tool to identify the reference time of each clause might help annotators and researchers identify \emph{Codas} and, consequently, \emph{Resolutions} more consistently.

However, this proposal has a limitation. \emph{Codas} can occur in the simple past tense; for example, ``And that was that’’ constitutes a \emph{Coda} \citep{Labov:2013}. When the verb appears in the simple past tense, the reference time coincides with the time of the past event. Therefore, in this case, the reference time is not shifted forward to the present. Reference time alone can serve only as a heuristic for identifying \emph{Codas}.

Another way to operationalize the concept of \emph{Coda} is to draw on the notion of narration time. \citet{NarrativeTime} distinguishes narration time from story time. Story time refers to the ``actual’’ temporal order of the narrated events. Narration time, in contrast, refers to the temporal perspective from which the story is presented to the reader or audience. A \emph{Coda} may be characterized as a clause in which narration time shifts from a perspective within story time to one outside it. Given the rapid advances in large language models in recent years, systems capable of reliably distinguishing narration time in narratives may become feasible and thus serve as tools for identifying \emph{Codas}.

\subsection{Reflections from Qualitative Researchers}

Interviews with qualitative researchers in philosophy and psychology helped us assess the usefulness of the annotation scheme beyond computational evaluation. A recurring pattern emerged regarding the relationship between affective valence and narrative structure.

Narratives about \emph{positive experiences}—such as feelings of happiness or satisfaction—were typically brief and self-contained, often forming a single story with relatively few \emph{Complications}. In contrast, accounts of \emph{difficulties or hardships} tended to be longer and more elaborated, often returning to earlier periods of struggle even after a \emph{Resolution} or \emph{Coda} had been narrated. Thus, closure markers did not always signal the end of the narrative trajectory; instead, speakers frequently returned to earlier events, further elaborating their accounts of hardship. This asymmetry may suggest that interviewees experienced difficulty in having their hardship listened to and understood by those around them.

This feedback highlights how researchers can derive insights by applying the Labovian model to oral narratives of personal experience.

\section{Conclusions and Future Work}
We presented the first systematic Labovian guidelines for annotating Japanese narratives.
The guidelines retain Labov’s six-category framework, add narrative types from \citet{Riessman:1990}, and provide clause segmentation rules for Japanese-specific constructions.
Clause segmentation by the three annotators showed consistently high agreement (Fleiss’ $\kappa = 0.80$), indicating that clause boundaries were stable and required little correction.
Structural-label agreement was moderate (Krippendorff’s $\alpha = 0.41$ and $0.45$, respectively), slightly higher than in previous studies despite finer-grained distinctions.
Future work will refine the guidelines, expand the dataset with more shareable data, and explore tools for scaling structural narrative analysis in Japanese.

\section{Ethics Statement}
The authors conducted all annotations in this study. 
All excerpts cited in the paper have been anonymized and modified to protect participants' privacy.
All interview participants provided informed consent before data collection. 
The study involves no sensitive personal data beyond the anonymized interview content, and we do not foresee any additional ethical risks arising from this research.
The study was approved by the Kobe University Ethics Committee (Ref: 2025-02).

\section{Data and Guideline Availability}
Due to privacy constraints, the interview transcripts used in this study cannot be made publicly available. 
However, we will release the latest version of the annotation guidelines developed in this project under a CC BY-SA 4.0 license. 
These guidelines, including decision charts and examples, are provided to facilitate collaboration and to support future research on Japanese narrative structure.

\section*{Acknowledgements}
This work was supported by JSPS KAKENHI Grant Number JP24H00809, JST CREST Grant Number JPMJCR2565, Japan. The author is grateful to Taisei Yamamoto, Ryoma Kumon, and Ruxiuan Tu for their valuable comments. Last, but not least, we thank Stephanie Nelson for generously sharing the final pre-publication draft of their article.

\nocite{*}
\section{Bibliographical References}\label{sec:reference}
\bibliographystyle{lrec2026-natbib}
\bibliography{lrec2026}

\section{Language Resource References}
\label{lr:ref}
\bibliographystylelanguageresource{lrec2026-natbib}
\bibliographylanguageresource{languageresource}

\end{document}

%% file: tables/dataset_example.tex
\begin{table*}[t]
\centering
\small
\renewcommand{\arraystretch}{1.2}
\setlength{\tabcolsep}{4pt}
\begin{tabularx}{\linewidth}{
  l
  X
  >{\centering\arraybackslash}m{3.6em}
  >{\centering\arraybackslash}m{3.6em}
  >{\centering\arraybackslash}m{6.8em}
}
\toprule
\textbf{\#} & \textbf{Caption} & \textbf{Story} & \textbf{Micro} & \textbf{Macro} \\
\midrule
1 & When I finally have time for myself, that is when I feel most fulfilled. & S & F & Abstract \\
2 & Last month I went out to a small mountain town. &  & F & Orientation \\
3 & I bought a lunch at the station &  & N & Complication \\
4 & and took the train out there. &  & N & Complication \\
5 & The sky was bright blue, &  & F & Orientation \\
6 & and I felt completely free. &  & R & Evaluation \\
7 & Sitting on the train, a cool breeze came in through the window. & & N & Complication \\
8 & I thought, ``Ah, autumn has come already.'' & & N & Resolution \\
9 & I hadn’t noticed the season passing at all this year. & & F & Orientation \\
10 & It was such a refreshing day, and I felt happy. & E & F & Coda \\
\bottomrule
\end{tabularx}
\caption{This table presents an example of clause segmentation and functional classification, based on an anonymized and extensively modified interview transcript translated from Japanese. Each clause is annotated with narrative span boundaries, as well as micro- and macro-level functional labels. \textbf{Story:} S/E indicates the starting and ending points of the narrative of the given type (Story, in this case). Columns corresponding to Habitual and Hypothetical narratives are omitted from this table for brevity. \textbf{Micro labels:} N = Narrative, R = Restricted, F = Free. This excerpt was produced in response to a question asking when the speaker feels a sense of satisfaction or happiness.}
\label{tab:dataset-example}
\end{table*}

%% file: tables/macro_labels_total.tex
\begin{table*}[tb]
\centering
\small
\renewcommand{\arraystretch}{1.2}
\setlength{\tabcolsep}{5pt}
\begin{tabularx}{\linewidth}{
  l
  X
  c
}
\toprule
\textbf{Label} & \textbf{Definition} & \textbf{Total} \\
\midrule
\textbf{Abstract} &
A short preview indicating the content or point of the story. Abstracts are often missing, like Codas, but when present they introduce the upcoming narrative in advance. &
17 \\
\addlinespace[2pt]
\textbf{Orientation} &
Introduces characters, time, place, or situation. Orientations guide the listener from the present moment into the past scene in which the story takes place. &
134 \\
\addlinespace[2pt]
\textbf{Complication} &
Reports ``what happened''—the sequence of past events that forms the core of the story. Complications may be interrupted by Evaluation or Orientation clauses. &
172 \\
\addlinespace[2pt]
\textbf{Evaluation} &
Provides the narrator’s stance, emotions, or interpretation of events, emphasizing why the story is worth telling. &
42 \\
\addlinespace[2pt]
\textbf{Resolution} &
Describes the outcome that concludes the chain of Complications. &
16 \\
\addlinespace[2pt]
\textbf{Coda} &
Returns the narrative to the present and signals its end, often explaining how the reported events continue to affect the present. &
18 \\
\bottomrule
\end{tabularx}
\caption{Macro-structural labels with total clause counts.}
\label{tab:macro-labels-total}
\end{table*}

%% file: tables/narrative_span_length.tex
\begin{table*}[tb]
\centering
\small
\renewcommand{\arraystretch}{1.2}
\setlength{\tabcolsep}{5pt}
\begin{tabularx}{\linewidth}{
  l
  X
  c
  c
}
\toprule
\textbf{Label} & \textbf{Definition} & \textbf{Narrative span} & \textbf{Mean length} \\
\midrule
\textbf{Story} &
A sequence of events that happened once at a specific time in the past. &
17 & 15.41 \\
\addlinespace[2pt]
\textbf{Habitual} &
Describes repeated or customary actions, routines, or recurring events. &
16 & 9.31 \\
\addlinespace[2pt]
\textbf{Hypothetical} &
Describes imagined or counterfactual events that did not actually occur. &
1 & 6.00 \\
\bottomrule
\end{tabularx}
\caption{Narrative span categories with definitions, number of spans, and mean clause counts with in a span in our dataset.}
\label{tab:narrative-span-length}
\end{table*}

%% file: tables/exact-match_micro.tex
\renewcommand{\arraystretch}{1.2}
\begin{tabular}{|c|c|}
\hline
\textbf{Micro label} & \textbf{Rate} \\
\hline
Narrative (N)   & \textbf{0.52} \\
\hline
Free (F)        & 0.15 \\
\hline
Restricted (R)  & 0.08 \\
\hline
\end{tabular}

%% file: tables/exact-match_macro.tex
\renewcommand{\arraystretch}{1.2}
\begin{tabular}{|c|c|}
\hline
\textbf{Macro label} & \textbf{Rate} \\
\hline
Abstract      & 0.13 \\
\hline
Orientation   & 0.30 \\
\hline
Complication  & \textbf{0.45} \\
\hline
Resolution    & 0.10 \\
\hline
Evaluation    & 0.11 \\
\hline
Coda          & 0.08 \\
\hline
\end{tabular}

%% file: main.bbl
\begin{thebibliography}{1}
\expandafter\ifx\csname natexlab\endcsname\relax\def\natexlab#1{#1}\fi

\bibitem[{Sald{\'\i}as and Roy(2020)}]{SaldiasRoy2020_RTNData}
Sald{\'\i}as and Roy. 2020.
\newblock \emph{{RTN:} Personal Narrative Dataset (Sald{\'\i}as \& Roy)}.
\newblock MIT-CCC. {MIT-CCC / ACL-NUSE Personal Narratives project, distributed via GitHub}.
\newblock PID \href{https://github.com/mit-ccc/acl-nuse-personal-narratives/blob/master/data/Saldias\%26Roy-RTN\_data.csv}{https://github.com/mit-ccc/acl-nuse-personal-narratives/blob/master/data/Saldias\%26Roy-RTN\_data.csv}.

\end{thebibliography}


\begin{thebibliography}{27}
\expandafter\ifx\csname natexlab\endcsname\relax\def\natexlab#1{#1}\fi

\bibitem[{Doi and Yanaka(2024)}]{doi_yanaka_2024_narrative_nlp}
Tomoki Doi and Hitomi Yanaka. 2024.
\newblock \href {https://doi.org/10.11517/jjsai.39.5_608} {Shizen gengo shori o mochiita naratibu bunseki no kanousei [possibilities of narrative analysis using natural language processing]}.
\newblock \emph{\textit{Journal of the Japanese Society for Artificial Intelligence}}, \textit{39}(\textit{5}):608--614.

\bibitem[{Fleischman(1990)}]{Fleischman:1990}
Suzanne Fleischman. 1990.
\newblock \emph{Tense and Narrativity}.
\newblock Routledge, London.

\bibitem[{Fournier(2013{\natexlab{a}})}]{Fournier:2013}
Chris Fournier. 2013{\natexlab{a}}.
\newblock \href {https://aclanthology.org/P13-1167} {Evaluating text segmentation using boundary edit distance}.
\newblock In \emph{Proceedings of the 51st Annual Meeting of the Association for Computational Linguistics (Volume 1: Long Papers)}, pages 1702--1712, Sofia, Bulgaria. Association for Computational Linguistics.

\bibitem[{Fournier and Inkpen(2012)}]{FournierInkpen:2012}
Chris Fournier and Diana Inkpen. 2012.
\newblock \href {https://aclanthology.org/N12-1016} {Segmentation similarity and agreement}.
\newblock In \emph{Proceedings of the 2012 Conference of the North American Chapter of the Association for Computational Linguistics: Human Language Technologies (NAACL-HLT 2012)}, pages 152--161, Montréal, Canada. Association for Computational Linguistics.

\bibitem[{Fournier(2013{\natexlab{b}})}]{Fournier2013-uv}
Christopher Fournier. 2013{\natexlab{b}}.
\newblock \emph{Evaluating Text Segmentation}.
\newblock Ph.D. thesis, Université d'Ottawa / University of Ottawa.

\bibitem[{Gordon and Swanson(2009)}]{GordonSwanson:2009}
Andrew~S. Gordon and Ron Swanson. 2009.
\newblock \href {https://www.icwsm.org/2009/data/Gordon-icwsm09-dcw.pdf} {Identifying personal stories in millions of weblog posts}.
\newblock In \emph{Proceedings of the International Conference on Weblogs and Social Media (ICWSM)}, pages \,\,\textit{pages unavailable}. Association for the Advancement of Artificial Intelligence.

\bibitem[{Kishimoto et~al.(2018)Kishimoto, Sawada, Murawaki, Kawahara, and Kurohashi}]{kishimoto-etal-2018-improving}
Yudai Kishimoto, Shinnosuke Sawada, Yugo Murawaki, Daisuke Kawahara, and Sadao Kurohashi. 2018.
\newblock \href {https://aclanthology.org/L18-1637/} {Improving crowdsourcing-based annotation of {J}apanese discourse relations}.
\newblock In \emph{Proceedings of the Eleventh International Conference on Language Resources and Evaluation ({LREC} 2018)}, Miyazaki, Japan. European Language Resources Association (ELRA).

\bibitem[{Kodama(2000)}]{Kodama:2000}
Yasue Kodama. 2000.
\newblock [{P}ossibilities and issues in analyzing {J}apanese narratives using the {L}abovian model] {R}abobian moderu ni yoru {N}ihongo no naratibu bunseki no kanōsei to shomondai (in {J}apanese).
\newblock \emph{Nihongo Kokusai Sent\={a} Kiy\={o}}, 10:17--32.

\bibitem[{Kubota et~al.(2024)Kubota, Sato, Amamoto, Akiyoshi, and Mineshima}]{kubota-etal-2024-annotation-japanese}
Ai~Kubota, Takuma Sato, Takayuki Amamoto, Ryota Akiyoshi, and Koji Mineshima. 2024.
\newblock \href {https://aclanthology.org/2024.lrec-main.109/} {Annotation of {J}apanese discourse relations focusing on concessive inferences}.
\newblock In \emph{Proceedings of the 2024 Joint International Conference on Computational Linguistics, Language Resources and Evaluation (LREC-COLING 2024)}, pages 1215--1224.

\bibitem[{Labov(2013)}]{Labov:2013}
William Labov. 2013.
\newblock \emph{The Language of Life and Death: The Transformation of Experience in Oral Narrative}.
\newblock Cambridge University Press, Cambridge, England.

\bibitem[{Labov and Waletzky(1967)}]{LabovWaletzky:1967}
William Labov and Joshua Waletzky. 1967.
\newblock Narrative analysis: Oral versions of personal experience.
\newblock In J.~Helm, editor, \emph{Essays on the Verbal and Visual Arts}, pages 3--38. University of Washington Press, Seattle and London.

\bibitem[{Lascarides and Asher(2007)}]{LascaridesAsher:2007}
Alex Lascarides and Nicholas Asher. 2007.
\newblock \emph{Segmentation and Interpretation in Discourse Representation Theory}.
\newblock Cambridge University Press.

\bibitem[{Levi et~al.(2022)Levi, Mor, Sheafer, and Shenhav}]{levi-etal-2022-detecting}
Effi Levi, Guy Mor, Tamir Sheafer, and Shaul~R. Shenhav. 2022.
\newblock \href {https://doi.org/10.18653/v1/2022.findings-naacl.133} {Detecting narrative elements in informational text}.
\newblock In \emph{Findings of the Association for Computational Linguistics: NAACL 2022}, pages 1755--1765, Seattle, United States. Association for Computational Linguistics.

\bibitem[{Mann and Thompson(1988)}]{MannThompson:1988}
William~C. Mann and Sandra~A. Thompson. 1988.
\newblock Rhetorical structure theory: Toward a functional theory of text organization.
\newblock \emph{Text}, 8(3):243--281.

\bibitem[{Mildner and Tamir(2024)}]{Mildner2024-spontaneous-thought}
Judith~N Mildner and Diana~I Tamir. 2024.
\newblock \href {https://doi.org/10.1093/pnasnexus/pgae230} {Why do we think? the dynamics of spontaneous thought reveal its functions}.
\newblock \emph{PNAS Nexus}, 3(6):pgae230.

\bibitem[{Nelson and Spence(2020)}]{NarrativeTime}
Stephanie Nelson and Barry Spence. 2020.
\newblock \href {https://doi.org/10.1093/acrefore/9780190201098.013.1076} {Narrative time}.

\bibitem[{Ouyang and McKeown(2014)}]{ouyang-mckeown-2014-towards}
Jessica Ouyang and Kathy McKeown. 2014.
\newblock \href {https://aclanthology.org/L14-1108/} {Towards automatic detection of narrative structure}.
\newblock In \emph{Proceedings of the Ninth International Conference on Language Resources and Evaluation ({LREC}'14)}, pages 4624--4631, Reykjavik, Iceland. European Language Resources Association (ELRA).

\bibitem[{Prasad et~al.(2008)Prasad, Dinesh, Lee, Miltsakaki, Robaldo, Joshi, and Webber}]{PennDiscourse:2008}
R~Prasad, N~Dinesh, Alan Lee, E~Miltsakaki, Livio Robaldo, A~Joshi, and B~Webber. 2008.
\newblock The penn discourse {TreeBank} 2.0.
\newblock \emph{LREC}, pages 2961--2968.

\bibitem[{Pustejovsky and Stubbs(2012)}]{Pustejovsky2012-xz}
James Pustejovsky and Amber Stubbs. 2012.
\newblock \emph{Natural language annotation for machine learning}.
\newblock O'Reilly Media, Inc.

\bibitem[{Rahimtoroghi et~al.(2014)Rahimtoroghi, Corcoran, Swanson, Walker, Sagae, and Gordon}]{Rahimtoroghi:2014}
Elahe Rahimtoroghi, Thomas Corcoran, Reid Swanson, Marilyn~A Walker, Kenji Sagae, and Andrew Gordon. 2014.
\newblock \href {https://aaai.org/papers/aaaiw-ws1310-14-9240/} {Minimal narrative annotation schemes and their applications}.
\newblock In \emph{Seventh Intelligent Narrative Technologies Workshop}. AAAI Publications.

\bibitem[{Reichenbach(1947)}]{Reichenbach:1947}
H.~Reichenbach. 1947.
\newblock \href {https://books.google.co.jp/books?id=xdA-AAAAIAAJ} {\emph{Elements of Symbolic Logic}}.
\newblock A Free Press paperback : philosophy. Macmillan Company.

\bibitem[{Riessman(1989)}]{Riessman:1989}
Catherine~Kohler Riessman. 1989.
\newblock Life events, meaning and narrative: the case of infidelity and divorce.
\newblock \emph{Social Science \& Medicine}, 29(6):743--751.

\bibitem[{Riessman(1990)}]{Riessman:1990}
Catherine~Kohler Riessman. 1990.
\newblock Strategic uses of narrative in the presentation of self and illness: a research note.
\newblock \emph{Social Science \& Medicine}, 30(11):1195--1200.

\bibitem[{Riessman(2008)}]{Riessman:2008}
Catherine~Kohler Riessman. 2008.
\newblock \emph{Narrative Methods for the Human Sciences}.
\newblock SAGE Publications, Thousand Oaks, CA.

\bibitem[{Sald\'ias and Roy(2020)}]{SaldiasRoy:2020}
Bel\'en Sald\'ias and Deb Roy. 2020.
\newblock \href {https://doi.org/10.18653/v1/2020.nuse-1.10} {Exploring aspects of similarity between spoken personal narratives by disentangling them into narrative clause types}.
\newblock In \emph{Proceedings of the First Joint Workshop on Narrative Understanding, Storylines, and Events}, pages 78--86, Online. Association for Computational Linguistics.

\bibitem[{Swanson et~al.(2014)Swanson, Rahimtoroghi, Corcoran, and Walker}]{swanson-etal-2014-identifying}
Reid Swanson, Elahe Rahimtoroghi, Thomas Corcoran, and Marilyn Walker. 2014.
\newblock \href {https://doi.org/10.3115/v1/W14-4323} {Identifying narrative clause types in personal stories}.
\newblock In \emph{Proceedings of the 15th Annual Meeting of the Special Interest Group on Discourse and Dialogue ({SIGDIAL})}, pages 171--180, Philadelphia, PA, U.S.A. Association for Computational Linguistics.

\bibitem[{Wasserscheidt et~al.(2021)Wasserscheidt, Mandić, Vollstädt, Jovanović, Tanasijević, Vučina~Simić, Yazhinova, and Zečević}]{Wasserscheidt2021-rb}
Philipp Wasserscheidt, Marija Mandić, Nadine Vollstädt, Ana Jovanović, Ivana Tanasijević, Ivana Vučina~Simić, Uliana Yazhinova, and Anđelka Zečević. 2021.
\newblock \href {http://dx.doi.org/10.22210/govor.2020.37.08} {Corpus‐based analysis of spoken narratives. introducing a corpus and a search tool}.
\newblock \emph{Govor/Speech}, 37(2):149--178.

\end{thebibliography}
